\documentclass[letterpaper]{article}
\usepackage{comment}
\usepackage{cite}
\usepackage{graphicx}
\usepackage{subfigure}
\usepackage{algorithm}
\usepackage{algpseudocode}
\algnewcommand{\LineComment}[1]{\State \(\triangleright\) #1}
\usepackage{aaai}
\usepackage{times}
\usepackage{helvet}
\usepackage{courier}
\usepackage{url}
\usepackage{color}
\usepackage{todonotes}
\usepackage{graphicx, subfigure}
\usepackage{amssymb}
\usepackage{epstopdf}
\usepackage{amsmath}
\usepackage{enumerate}
\usepackage{amsthm}
\usepackage{appendix}

\usepackage{times}
\usepackage{helvet}
\usepackage{courier}
\begin{document}
%
\title{The Boundary Forest Algorithm for Online Supervised and Unsupervised Learning}
\author{Charles Mathy\\
Disney Research \\
Boston\\
cmathy\\
@disneyresearch.com\\
\And
Nate Derbinsky\\
Wentworth Institute \\
of Technology\\
derbinskyn\\
@wit.edu\\
\And
Jos\'e Bento\\
Boston\\
 College\\
jose.bento\\
@bc.edu\\
\And
Jonathan Rosenthal\\
Disney Research \\
Boston\\
jon.rosenthal\\
@disneyresearch.com\\
\And
Jonathan Yedidia\\
Disney Research\\
 Boston\\
yedidia\\
@disneyresearch.com\\
}
\maketitle
\begin{abstract}
\begin{quote}
We describe a new instance-based learning algorithm called the Boundary Forest (BF) algorithm, that can be used for supervised and unsupervised learning. The algorithm builds a forest of trees whose nodes store previously seen examples. It can be shown data points one at a time and updates itself incrementally, hence it is naturally online. Few instance-based algorithms have this property while being simultaneously fast, which the BF is. This is crucial for applications where one needs to respond to input data in real time. The number of children of each node is not set beforehand but obtained from the training procedure, which makes the algorithm very flexible with regards to what data manifolds it can learn. We test its generalization performance and speed on a range of benchmark datasets and detail in which settings it outperforms the state of the art. Empirically we find that training time scales as $O( D N log(N) )$ and testing as $O( D log(N) )$, where $D$ is the dimensionality and $N$ the amount of data.
\end{quote}
\end{abstract}


\section{Introduction}
\label{sec:intro}

The ability to learn from large numbers of examples, where the examples themselves are often high-dimensional, is 
vital in many areas of machine learning. Clearly, the ability to generalize from training examples to test queries 
is a key feature
that any learning algorithm must have, but there are several other features that are also crucial in many practical situations.
In particular, we seek a learning algorithm that is: (i) fast to train, (ii) fast to query, 
(iii) able to deal with arbitrary data distributions, and (iv) able to learn incrementally in an online setting. 
Algorithms that satisfy all these properties, particularly (iv), are hard to come by, however they are of immediate importance in problems 
such as real time computer vision, robotic control, and more generally, problems which involve learning from and
responding quickly to streaming data.

We present here the Boundary Forest (BF) algorithm that satisfies all these properties, and as a bonus, is 
transparent and easy to implement.
The data structure underlying the BF algorithm is a collection of boundary trees (BTs). The nodes
in a BT each store a training example. 
The BT structure can be efficiently queried at query time and quickly modified to incorporate new data points during training. 
The word ``boundary'' in the name 
relates to its use in classification, where most of the nodes in a BT will be near the boundary between different classes. 
The method is nonparametric and can learn arbitrarily shaped boundaries, as the tree structure is determined from the data and
not fixed {\em a priori}.
The BF algorithm is very flexible; in essentially the same form, it can be used for classification, regression and nearest
neighbor retrieval problems.


\section{Related work} \label{sec:back}

There are several existing methods, including KD-trees \cite{Friedman_1977}, Geometric Near-neighbor Access Trees \cite{Brin_1995}, and Nearest Vector trees \cite{Lesjek_2011} that build tree search structures on large datasets (see \cite{Samet_2006} for an extensive bibliography). These algorithms typically need batch access to the entire dataset before constructing their trees, in which case they may outperform the BF, however we are interested in an online setting. Two well known tree-based algorithms that allow online insertion are cover trees \cite{beygelzimer2006covertree} and ball trees. The ball tree online insertion algorithm \cite{omohundro1989five} is rather costly, requiring a volume minimizing step at each addition. The cover tree, on the other, has a cheap online insertion algorithm, and it comes with guarantees of query time scaling as $O(c^6 log N)$ where N is the amount of data and c the so-called expansion constant, which is related to the intrinsic dimensionality of the data. We will compare to cover trees below. Note that $c$ in fact depends on $N$ as it is defined as a worse case computation over the data set. It can also diverge from adding a single point. 

Tree-based methods can be divided into those that rely on calculating metric distances between points to move down the tree, and those that perform cheaper computations. Examples of the former include cover trees , ball trees and the BF algorithm we present here. Examples of the latter include random decision forests (RFs) and kd trees  \cite{Friedman_1977} . In cases where it is hard to find a useful subset of informative features, metric-based methods may give better results, otherwise it is of course preferable to make decisions with fewer features as this makes traversing the trees cheaper. Like other metric-based methods, the BF can immediately be combined with random projections to obtain speedup, as it is known by the Johnson-Lindenstrauss lemma \cite{johnsonlindenstrauss1987} that the number of projections needed to maintain metric distances only grows as $log(D)$ as the data dimensionality $D$ grows. 
There has been work on creating online versions of RFs \cite{Kalal_2009} and kd trees. In fact, kd trees typically scale no better than brute force in higher than 20 dimensions \cite{muja2009fast} , but multiple random kd trees have been shown to overcome this difficulty. We will compare to offline RFs and online random kd trees (the latter implemented in the highly optimized library FLANN \cite{silpa2008optimised}) below.





The naive nearest neighbor algorithm is online, and there is extensive work on trying to reduce the number of stored nearest neighbors to 
reduce space and time requirements \cite{Aha_1991}. As we will show later, we can use some of these methods in our approach. 
In particular, for classification the Condensed Nearest Neighbor algorithm \cite{Wilson_2000} only adds a point if the 
previously seen points misclassify it. This allows for a compression of the data and significantly accelerates learning, and we use the same idea
in our method.
Previous algorithms that generate a tree search structure would have a hard time doing this, as they need enough data from the outset to build the tree. 


\section{The Boundary Forest algorithm} \label{sec:co}

A boundary forest is a collection of $n_T$ rooted trees. Each tree consists of nodes representing training examples, with edges between nodes created
during training as described below. The root node of each tree is the starting point for all queries using that tree.
Each tree is shown a training example or queried at test time
independently of the other trees; thus one can trivially parallelize training and querying.

Each example has a $D$-dimensional real position $x$ and a ``label'' vector $c(x)$ associated with it (for retrieval problems, one can think of
$c(x)$ as being equal to $x$, as we will explain below). For example, if one is dealing with a 10-class classification problem, we could 
associate a 10-dimensional indicator vector $c(x)$ with each point $x$.

One must specify a metric associated with the positions $x$, 
which takes two data points $x$ and $y$ and outputs a real number $d(x,y)$. Note that in fact this ``metric'' can be any real function, 
as we do not use any metric properties, but for the purpose of this paper we always use a proper metric function. 
Another parameter that one needs to specify is an integer $k$ which represents the maximum number of child nodes connected to any node in the tree.

Given a query point $y$, and a boundary tree $T$, the algorithm moves through the tree starting from the root node, and 
recursively compares the
distance to the query point from the current node and from its children, moving to and recursing at 
the child node that is closest to the query, unless the
current node is closest and has fewer children than $k$, in which case it returns the current node.  
This greedy procedure finds a ``locally closest'' example to the query, in the sense that none of the children of the locally closest node are closer. Note that the algorithm is not allowed to stop at a point that already has $k$ children, because it could potentially get a new child if the current training point is added. As we will show, having finite $k$ can significantly improve speed at low or negligible cost in performance.

\begin{algorithm}
\caption{The Boundary Tree (BT) algorithm}
\textbf{associated data} \\
\quad 	rooted tree - the $i$th node has position $x_i$ and label vector $c(x_i)$ \\
\quad 	$\epsilon$ the real threshold for comparison of label vectors\\
\quad 	$d$ the metric for comparing positions\\
\quad 	$d_c$ the metric for comparing label vectors\\
\quad 	$k$ the maximum number of children per node ($k>1$)\\
\begin{algorithmic}[1]
\Procedure {BTQuery}{$y$} \\
\textbf{input} $y$ the query position\\
\textbf{output} node $v$ in Boundary Tree.\\
\textbf{start} initialize $v$ at the root node $v_0$\\
\textbf{do} \\
\quad	define $A_v$ to be the set of nodes consisting of the children of $v$ in the Boundary Tree\\
\quad	\textbf{if} the number of children of $v$ is smaller than $k$ \textbf{then} add $v$ to $A_v$ \textbf{end if} \\
\quad	let $v_{min} = \textrm{argmin}_{w\in A_v} d(w,x)$, i.e. the node in $A_v$ that is closest to $y$	(choose randomly from any ties) \\
\quad	\textbf{if} $v_{min}=v$ \textbf{then} break \textbf{end if}\\
\quad	$v \gets v_{min}$ \\
\textbf{end do}	\\
\textbf{return} $v$ 
\EndProcedure
\Statex
\Procedure{BTTrain}{$y$,$c(y)$} \\
\textbf{input} \\
\quad position $y$\\
\quad label vector $c(y)$\\
\textbf{start} $v_{min} = \Call{BoundaryTreeQuery}{y}$\\
\textbf{if $d_c(c(y),c(v_{min}))>\epsilon$} \textbf{then}\\
\quad create node $v_{new}$ in Boundary Tree with position $y$ and label $c(y)$\\
\quad add edge from $v_{min}$ to $v_{new}$\\
\textbf{end if} 
\EndProcedure
\end{algorithmic}
\label{alg:BT}
\end{algorithm}

\begin{algorithm}
\caption{The Boundary Forest (BF) algorithm}
(see algorithm \ref{alg:BT} for definition of subroutines called here)
\textbf{associated data} \\
\quad Forest of $n_T$ BTs: $BF = \{ BT_1,\ldots,BT_{n_T}\}$\\
\quad all BTs have same associated $d$, $d_c$, $\epsilon$, $k$ \\
\quad $E$ the estimator function that takes in a position $x$, a set of $n_T$ nodes $v_i$ consisting of positions and label vectors, and outputs a label vector $E(x,v_1,\ldots,v_{n_T})$ \\
\textbf{initialization} \\
\quad start with $n_T$ training points, at positions $y_1,\ldots,y_{n_T}$ and with respective labels $c(y_1),\ldots,c(y_{n_T})$.\\
\quad Call $\Call{BFInitialization}{y_1,c(y_1),\ldots,y_{n_T},c_{n_T}}$\\
\begin{algorithmic}[1]
\Procedure{BFQuery}{$y$} \\
\textbf{input} position $y$\\
\textbf{start} \\
\textbf{for $i$ from 1 to $n_T$}\\
\quad $v_i=BT_i.\Call{BTQuery}{y}$\\
\textbf{end for} \\
\textbf{return} $E(y,v_1,\ldots,v_{n_T})$
\EndProcedure
\Statex
\Procedure{BFTrain}{$y$,$c(y)$} \\
\textbf{input} \\
\quad position $y$\\
\quad label vector $c(y)$\\
\textbf{start} \\
\textbf{for $i$ from 1 to $n_T$}\\
\quad call $BT_i.\Call{BTTrain}{y,c(y)}$ \\
\textbf{end for}  
\EndProcedure
\Statex
\Procedure{BFInitialization}{$y_1$,$c(y_1)$,$\ldots$,$y_{n_T}$,$c_{n_T}$} \\
\textbf{input} \\
\quad positions $y_1,\ldots,y_{n_T}$\\
\quad label vectors $c(y_1),\ldots,c(y_{n_T})$\\
\textbf{start} \\
\textbf{for $i$ from 1 to $n_T$}\\
\quad set position of root node of $BT_i$ to be $y_i$, and its label vector $c(y_i)$ \\
\quad \textbf{for $j$ from 1 to $n_T$} \\
\quad \quad \textbf{if $i\neq j$} \textbf{then } call $BT_i.\Call{BTTrain}{y_j,c(y_j)}$\\
\quad \textbf{end for}\\
\textbf{end for}  
\EndProcedure
\end{algorithmic}
\label{alg:BF}
\end{algorithm}



Once each tree has processed a query, one is left with a set of $n_T$ locally closest nodes. 
What happens next depends on the task at hand: if we are interested in retrieval, then we take the closest of those $n_T$ locally closest 
nodes to $y$ as the 
approximate nearest neighbor. For regression, given the positions $x_i$ of the locally closest nodes and their associated vector $c(x_i)$ 
one must combine them to form an estimate for $c(y)$. Many options exist, but in this paper we use a Shepard weighted average \cite{shepard1968two}, where the
weights are the inverse distances, so that the estimate is 
\begin{equation}
c(y) = \frac{\sum_i c(x_i)/d(x_i,y)} { \sum_i 1/d(x_i, y)}.
\end{equation}
For classification, as described above we use an indicator function $c(x)$ for the training points. Our answer is the class corresponding
to the coordinate with the largest value from $c(y)$; we again use Shepard's method to determine $c(y)$. Thus, for a three-class problem where the
computed $c(y)$ was $[0.5, 0.3, 0.2]$, we would return the first class as the estimate. For regression, the output of the BF is simply the the Shepard weighted average of the locally closest nodes $x_i$ output by each tree. Finally, for retrieval we take the  node $x^*$ of the locally closest nodes $x_i$ that is closest to the query $y$.

Given a training example with position $z$ and ``label'' vector $c(z)$, we first query each tree in the forest using $z$ as we just described. 
Each tree independently outputs a locally closest node $x_i$ and decides whether a new node should be created with associated position $z$ and label $c(z)$ 
and connected by an edge to $x_i$. The decision depends once again on the task: for classification, the node is created if $c(x_i) \neq c(z)$. 
For regression, one has to define a threshold $\epsilon$ and create the node if $|(c(x_i)-c(z)|>\epsilon$. Intuitively, the example is added to a tree
if and only if the current tree's prediction of the label was wrong and needs to be corrected.
For retrieval, we add all examples to the trees.

If all BTs in a BF were to follow the exact same procedure, they would all be the same. To decorrelate the trees, 
the simplest procedure is to give them each a different root. Consider $BF=\{BT_1,\ldots,BT_{n_T}\}$, a BF of $n_T$ trees. 
What we do in practice is take the first $n_T$ points, make point $i$ the root node of $BT_i$, and use as the other $n_T-1$ initial
training points for each $BT_i$ a 
random shuffling of the remaining $n_T-1$ nodes. We emphasize that after the first $n_T$ training nodes (which are a very small fraction
of the examples) the algorithm is strictly online.

For the algorithm just described, we find empirically that query time scales as a power law in the amount of data $N$ with a power $\alpha$ smaller than $2$, which implies that training time scales as $N^{1+\alpha}$ (since training time is the integral of query time over $N$).
We can get much better scaling by adding a simple change: we set a maximum $k$ of the number of children a node in the BF can have. The algorithm cannot stop at a node with $k$ children. With this change, query time scales as $log(N)$ and training time as $Nlog(N)$, and if $k$ is large enough performance is not negatively impacted. The memory scales linearly with the number  of nodes added, which is linear in the amount of data $N$ for retrieval, and typically sublinear for classification or regression as points are only added if misclassified. We only store pointers to data in each tree, thus the main space requirement comes from a single copy of each stored data point and does not grow with $n_T$.

The BF algorithm has a very appealing property which we call \emph{immediate one-shot learning}: if it is shown a training example and immediately afterwards queried at that point, it will get the answer right. In practice, we find that the algorithm gets zero or a very small error on the training set after one pass through it (less than $1\%$ for all data sets below).

A pseudo-code summary of the algorithm for building Boundary Trees and Boundary Forests is given in Algorithms \ref{alg:BT} and \ref{alg:BF} respectively.

\section{Scaling properties}

To study the scaling properties of the BF algorithm, we now focus on its use for retrieval. Consider examples drawn uniformly from within a 
hypercube in $D$ dimensions. The qualitative results we will discuss are general: we tested a mixture of Gaussians of arbitrary 
size and orientation, and real datasets such as MNIST treated as a retrieval problem, and obtained the same picture. We will 
show results for a uniformly sampled hypercube and unlabeled MNIST. Note that we interpret raw pixel intensity values as vectors for MNIST without
any preprocessing, and throughout this paper the Euclidean metric is used for all data sets. 
We will be presenting scaling fits to the different lines, ruling out one scaling law over another. In those cases, our procedure was to take the first half of the data points, fit them separately to one scaling law and to the one we are trying to rule out, and look at the rms error over the whole line for both fits. The fits we present have rms error at least 5 times smaller than the ruled out fits.

\begin{figure}[h!]
\centering
\includegraphics[width=0.23\textwidth]{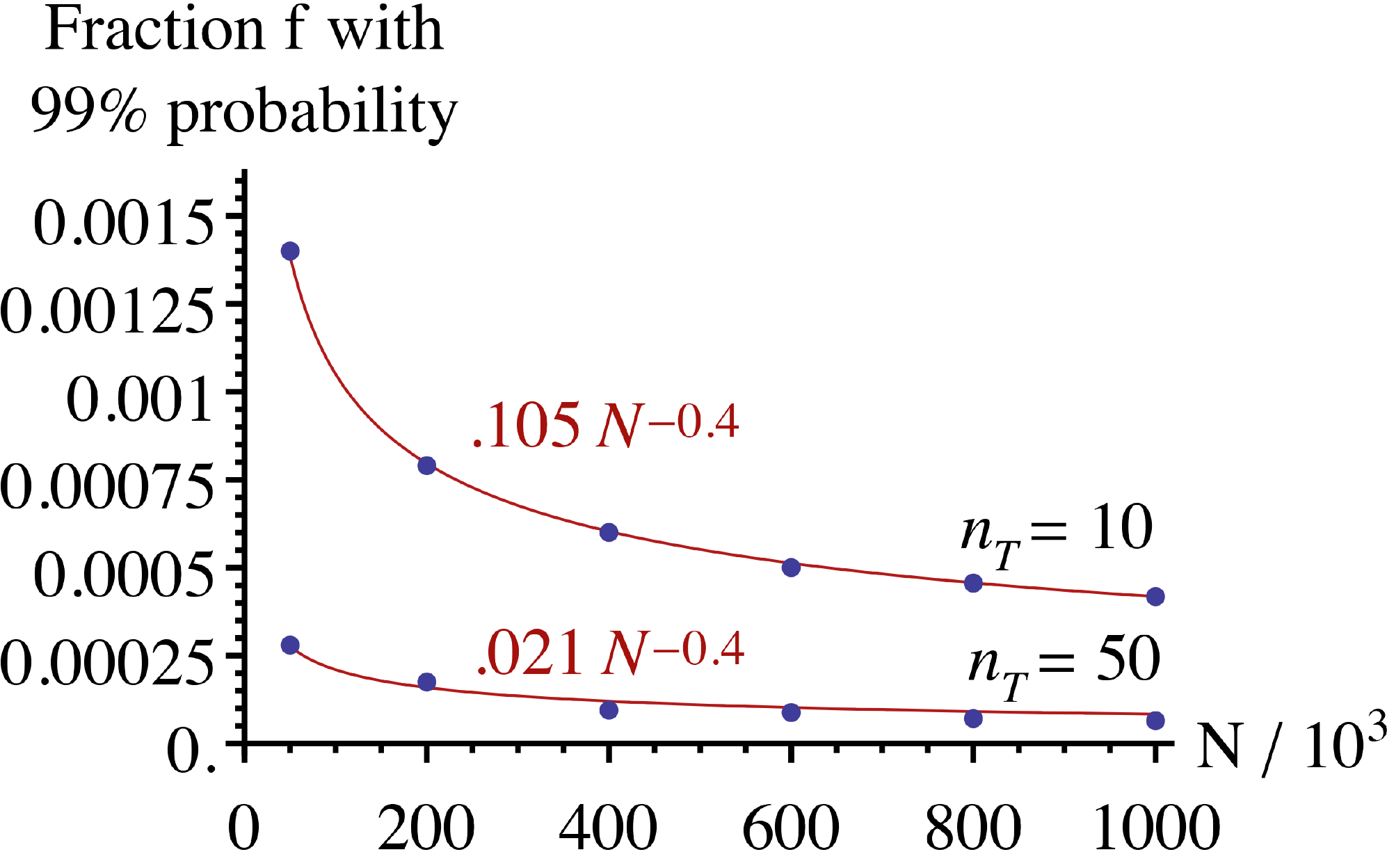}
\includegraphics[width=0.23\textwidth]{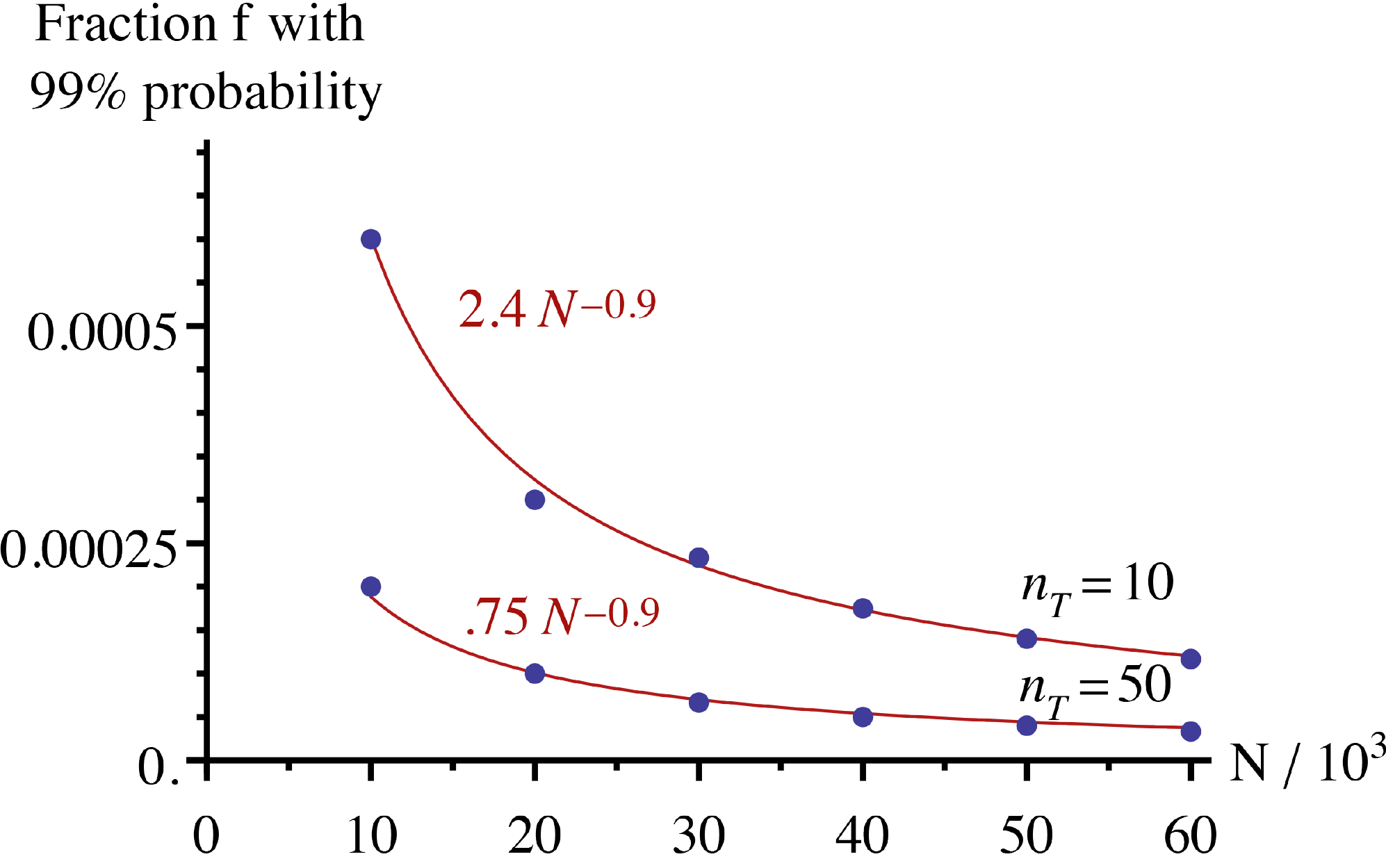}
\put(-245,65){\tiny (a)}
\put(-125,65){\tiny (b)}
\caption{Retrieval accuracy for a BF with $k=50$ and $n_T=10$ or $50$ receiving data from (a) a uniformly sampled 100 dimensional hypercube, (b) MNIST.
The x-axis gives the number of training examples $N$. 
The y-axis shows the fraction $f$ such that the BF has $99\%$ chance of outputting an example that is 
within the $f N$ closest training examples to the query. Shown are a few plots for different values of the number of trees $n_T$, and the maximum number of children $k$ per node, in the $BF$. The retrieval accuracy improves as a power law with $N$, with a power that depends on the data set. }
\label{fig:retrieval}
\end{figure}

Denote by $N$ the number of training examples shown to a BF algorithm using $n_T$ trees and a maximum number of children per 
node $k$. Recall that for retrieval on a query point $y$, the BF algorithm returns a training example $x^*$ which is the closest of
the locally closest nodes from each tree to the query. To assess the performance, 
we take all training examples, order them according to their distance from $y$, 
and ask where $x^*$ falls in this ordering. We say $x^*$ is in the $f$ best fraction 
if it is among the $fN$ closest points to $y$. 
In Fig. \ref{fig:retrieval} we show the fraction $f$ obtained if we require $99\%$ probability that $x^*$ is within the $fN$ closest examples to $y$. 
We see that the fraction $f$ approaches zero as a power law as the number of training examples $N$ increases. 

\begin{figure}[h!]
\centering
\includegraphics[width=0.23\textwidth]{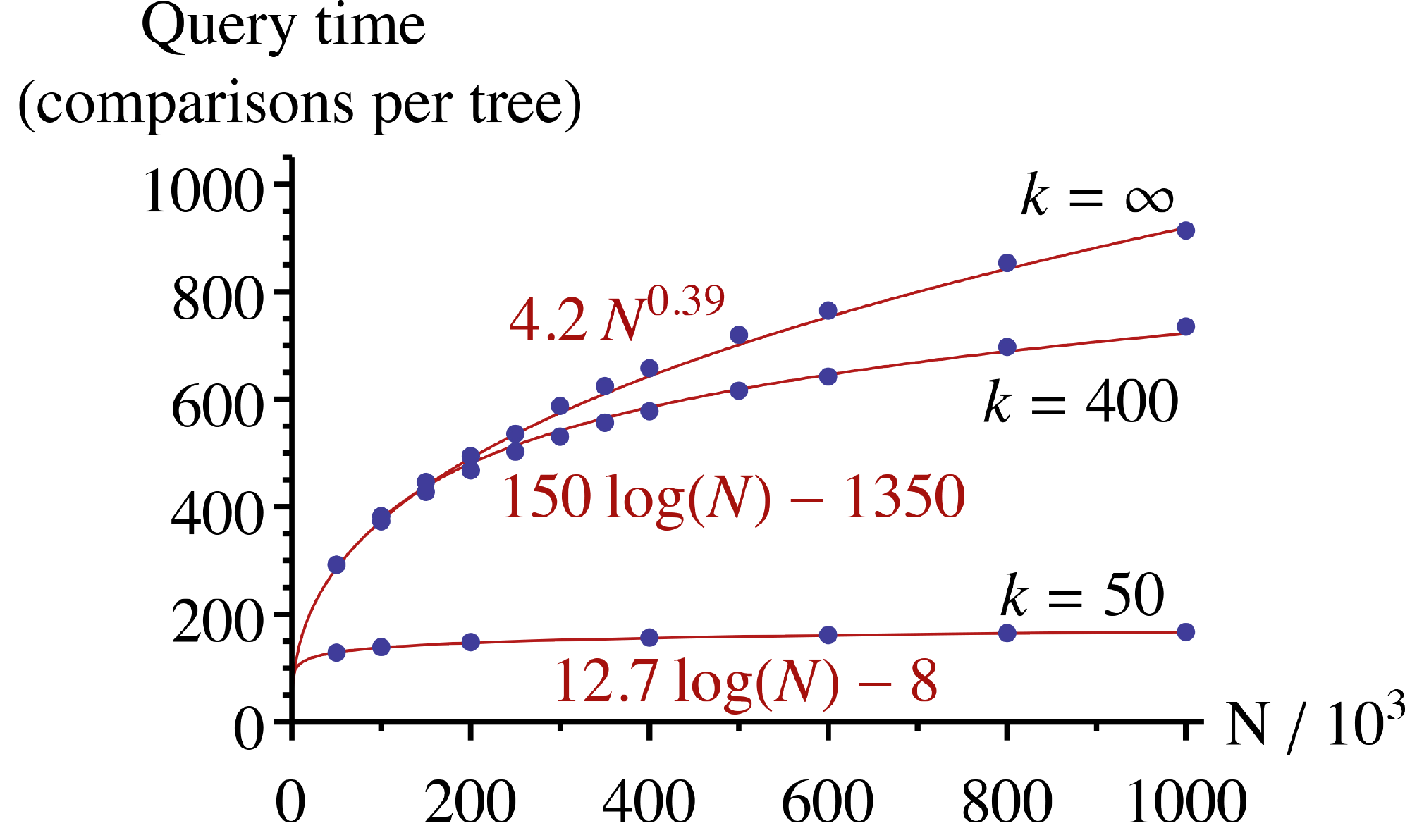}
\includegraphics[width=0.23\textwidth]{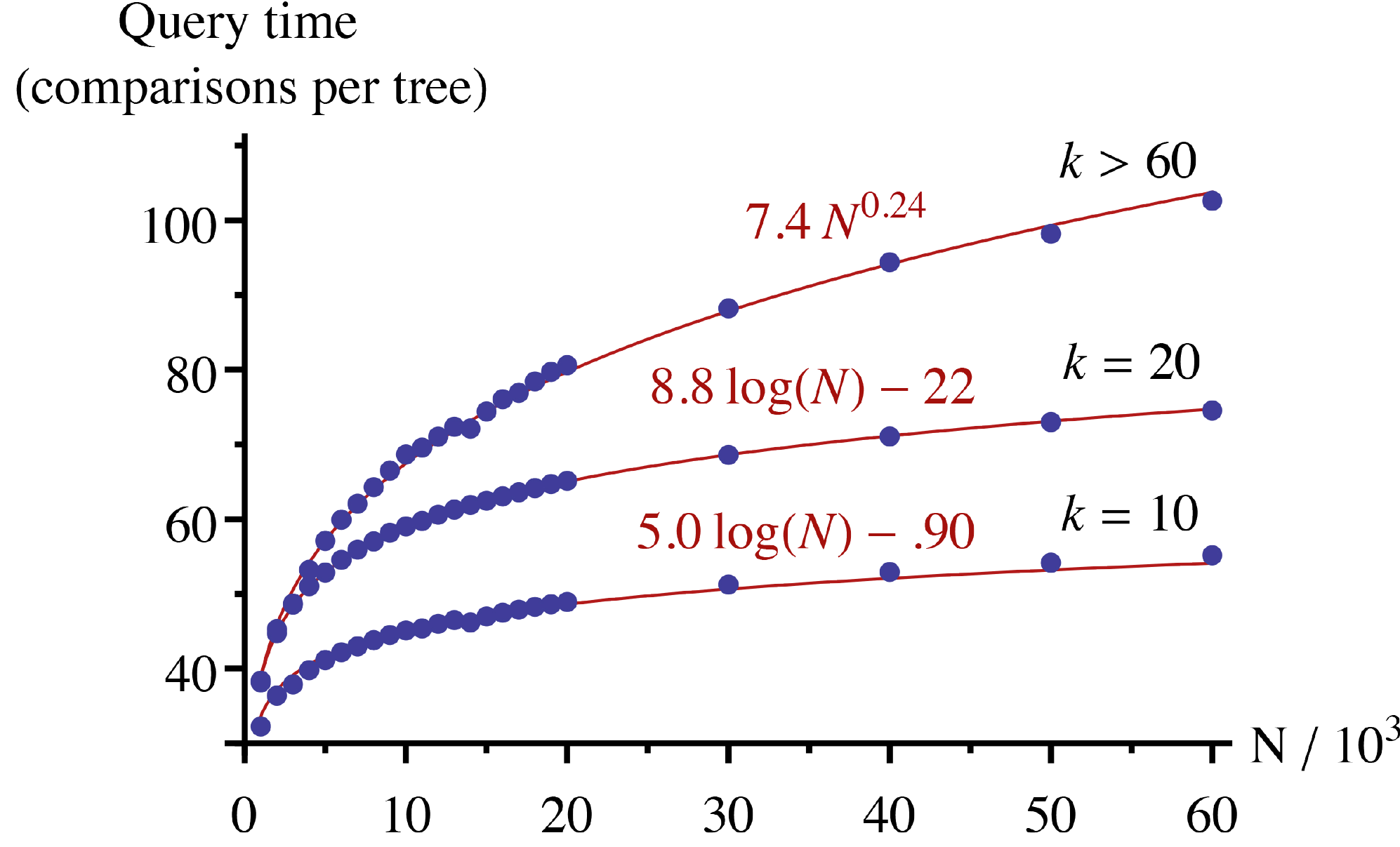}
\put(-240,70){\tiny (a)}
\put(-120,70){\tiny (b)}
\caption{Query time per tree of a BF with ten trees after having been exposed to $N$ examples, in units of metric comparisons between the query and nodes 
per tree, for two datasets: (a) data uniformly distributed in the 100 dimensional hypercube; (b) MNIST treated as a retrieval problem. 
We observe that once the root node has $k$ children, the scaling changes from being power law to logarithmic. 
Note that where we indicate logarithmic scaling, this scaling is only valid after the line departs from the $k=\infty$ line.
The power law when $k=\infty$ depends on the data set. 
For MNIST, $k > 60$ is equivalent to $k=\infty$, because no node hits the $k=60$ limit. 
If the data set were larger, for any finite $k$ one would eventually see the behavior switch from power law to logarithmic. The results vary very little
for different BFs obtained using shuffled data.}
\label{fig:querytimes}
\end{figure}

Next we consider the query time of the BF as a function of the number of examples $N$ it has seen so far. Note that 
training and query time are not independent: since training involves a query of each BT in the BF, followed by adding the node to the
 BT which takes negligible time, training time is the integral of query time over $N$. In Fig. \ref{fig:querytimes} we plot the query time (measured
in numbers of metric comparisons per tree, as this is the computational bottleneck for the BF algorithm) 
as a function of $N$, for examples drawn randomly from the 100-dimensional hypercube and for unlabeled MNIST. What we observe is that if $k=\infty$, query time scales sublinearly in $N$, with a power that depends on the dataset, but smaller than $0.5$. However, for finite $k$, the scaling is initially sublinear but then it switches to logarithmic. This switch happens around the time when nodes in the BT start appearing with the number of children equal to $k$. 

To understand what is going on, we consider an artificial situation where all points in the space are equidistant, which removes any 
 from the problem. In this case, we once again have a root node where we start, and we will go from a node to one of its children recursively until we stop, at which point we connect a new node to the node we stopped at. The rule for traversal is as follows: if a node has $q$ children, then with probability $1/(q+1)$ we stop at this node and connect a new node to it, while with probability $q/(q+1)$ we go down one of its children, all with equal probability $1/(q+1)$.

\begin{figure}[h!]
\centering
\includegraphics[width=0.25\textwidth]{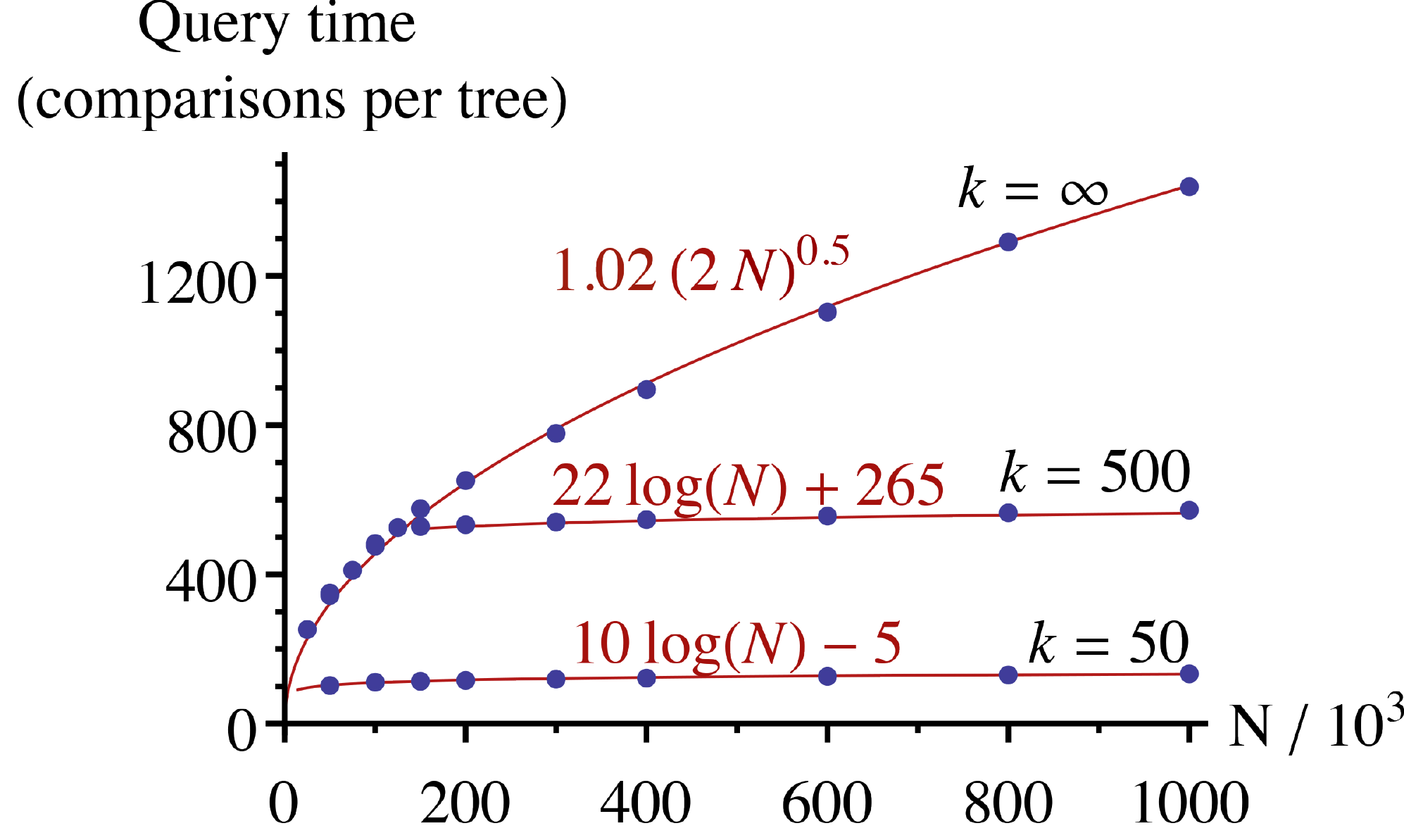}
\includegraphics[width=0.21\textwidth]{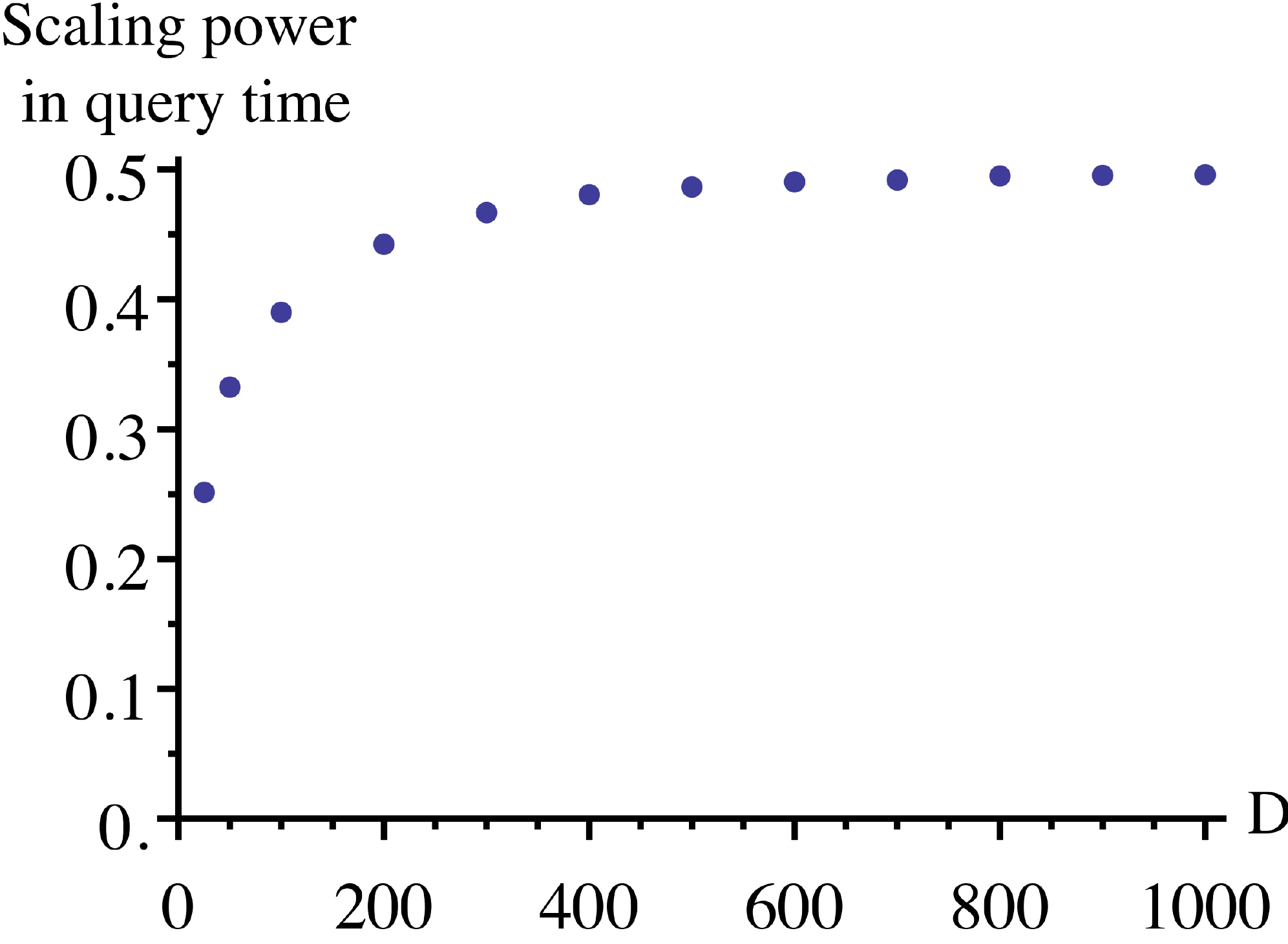}
\caption{
Scaling of the query time of the artificial BT described in the text, where the BF connects a training example to a node or goes down to one of its children all with equal probability. 
(a) shows the the query time of the artificial BT as a function of the amount of training examples $N$, for different values of the maximum number of children $k$. 
We see that for finite $k$, the scaling of query time is initially $N^{0.5}$ but eventually switches to logarithmic, at a time that grows with increasing $k$. Indeed, this time coincides with the point where the root has the maximum number $k$ of children.
In (b) the power law $\alpha$ of the scaling of query time for $k=\infty$, which then scales as $N^{\alpha}$, is shown as a function of data dimensionality $D$, for a BT trained on examples drawn uniformly from the $D$-dimensional unit hypercube. As $D\rightarrow\infty$, $\alpha\rightarrow 0.5$. We can understand the limit $D\rightarrow\infty$ by realizing in that limit the BT behaves like the
artificial BT.}
\label{fig:querytimesTT}
\end{figure}


The query time for this artificial tree is $ (2N)^{0.5}$ for large $N$ (plus subleading corrections), as shown in Fig. \ref{fig:querytimesTT} (a). To understand why, consider the root node. 
If it has $q-1$ children, the expected time to add a new child is $q$. Therefore the expected number of steps for the root node to have $q$ children scales as $q^2/2$.
Thus the number of children of the root node, and the number of metric comparisons made at the root grows as $\sqrt{2 N}$ (set $q = \sqrt{2N}$). 
We find that numerically the number of metric comparisons scales around $1.02 \sqrt{2N}$, which indicates that the metric comparisons to the root's children is the main computational cost. The reason is that the root's children have 
$\sim N^{1/4}$ children, as can be seen by repeating the previous scaling argument. 
If, on the other hand, we set $k$ to be finite, initially the tree will behave as though $k$ was infinite, until the root node has $k$ children, 
at which point it builds a tree where the query time grows logarithmically, as one would expect of 
an approximately balanced tree with $k$ children or less per node.

In data sets with a metric, we find the power law when $k=\infty$ to be smaller than $0.5$. Intuitively, this occurs because new children of the root must be closer to the root than any of its other children, therefore they reduce the probability that further query points are closer to the
root than its children. As the dimensionality $D$ increases, this effect diminishes, as new points have increasingly small inner products with each other, and if all points were orthogonal you do not see this bias. In Fig. \ref{fig:querytimesTT}(b) we plot the power $\alpha$ in the scaling $O(N^{\alpha})$ of the query time of a BT trained on data drawn uniformly from the $D$-dimensional hypercube, and find that as $D$ increases, $\alpha$ approaches $0.5$ from below, which is consistent with the phenomenon just described.

We now compare to the cover tree (CT) algorithm \footnote{We adapted the implementation of \cite{CTgithub} - which was the only implementation we could find of the online version of CT - to handle approximate nearest neighbor search.}. For fair comparison, we train the CT adding the points using online insertion, and when querying we use the CT as an approximate nearest neighbor (ANN) algorithm (to our knowledge, this version of the CT which is defined in \cite{beygelzimer2006covertree} has not been studied previously). In the ANN incarnation of CT, one has to define a parameter $\epsilon$, such that when queried with a point $p$ the $CT$ outputs a point $q$ it was trained on such that $d(p,q) \leq (1+\epsilon) d_{min}(p)$ where $d_{min}(p)$ is the distance to the closest point to $p$ that the CT was trained on. We set $\epsilon=10$ ($\epsilon$ has little effect on performance or speed: see Appendix for results with $\epsilon=0.1$).

Another important parameter is the base parameter $b$. It is set to $2$ in the original proofs, however the original creators suggest that a value smaller than $2$ empirically leads to better results, and their publicly available code has as a default the value $1.3$, which is the value we use. Changing this value can decrease training time at the cost of increasing testing time, however the scaling with amount of data remains qualitatively the same (see Appendix for results with $b=1.1$). Note that the cover tree algorithm is not scale invariant, while the BF is: if all the features in the data set are rescaled by the same parameter, the BF will do the exact same thing, which is not true of the CT. Also, for the CT the metric must satisfy the triangle inequality, and it is not obvious if it is parallelizable. 

\begin{figure}[h!]
\centering
\includegraphics[width=0.23\textwidth]{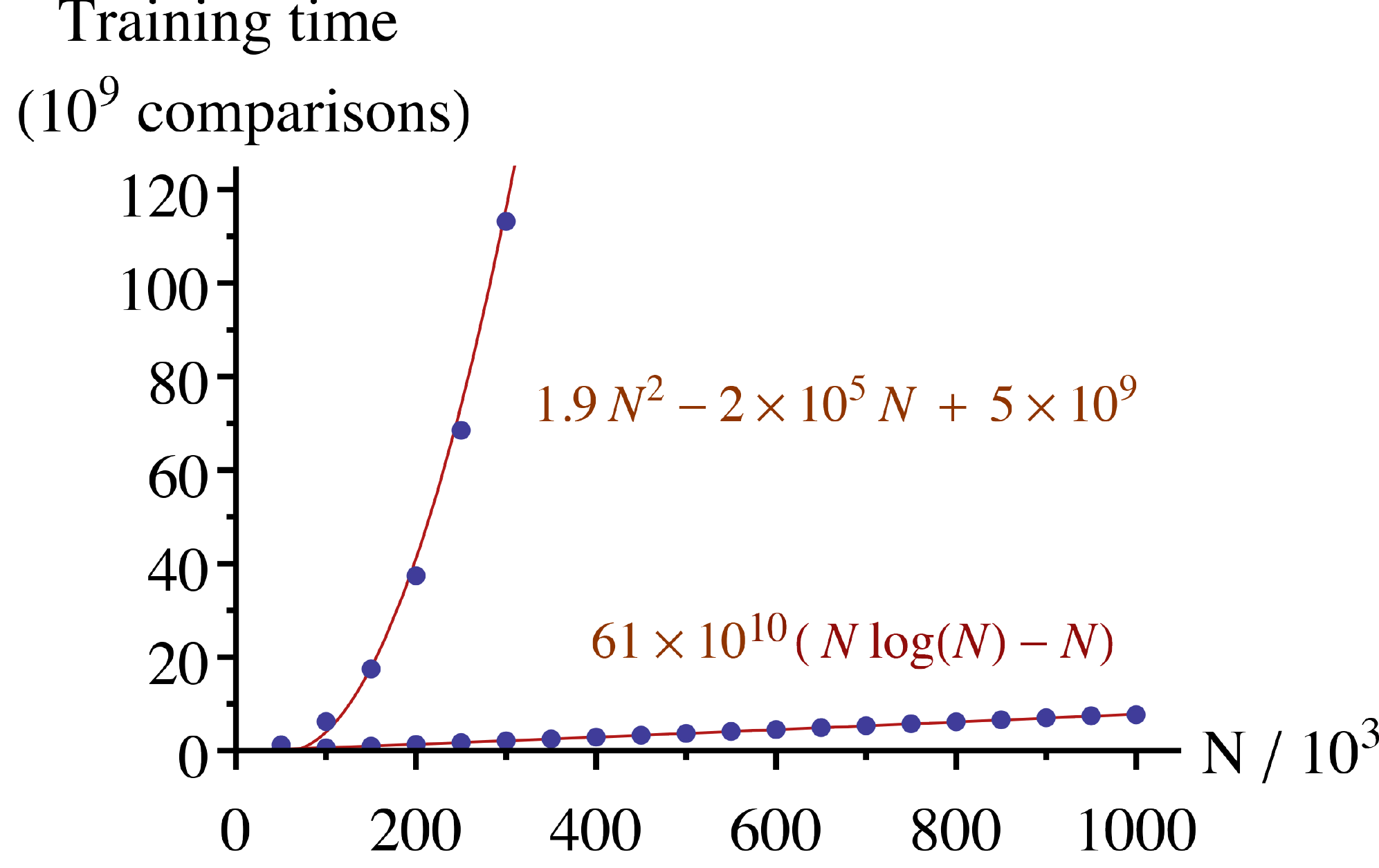}
\includegraphics[width=0.23\textwidth]{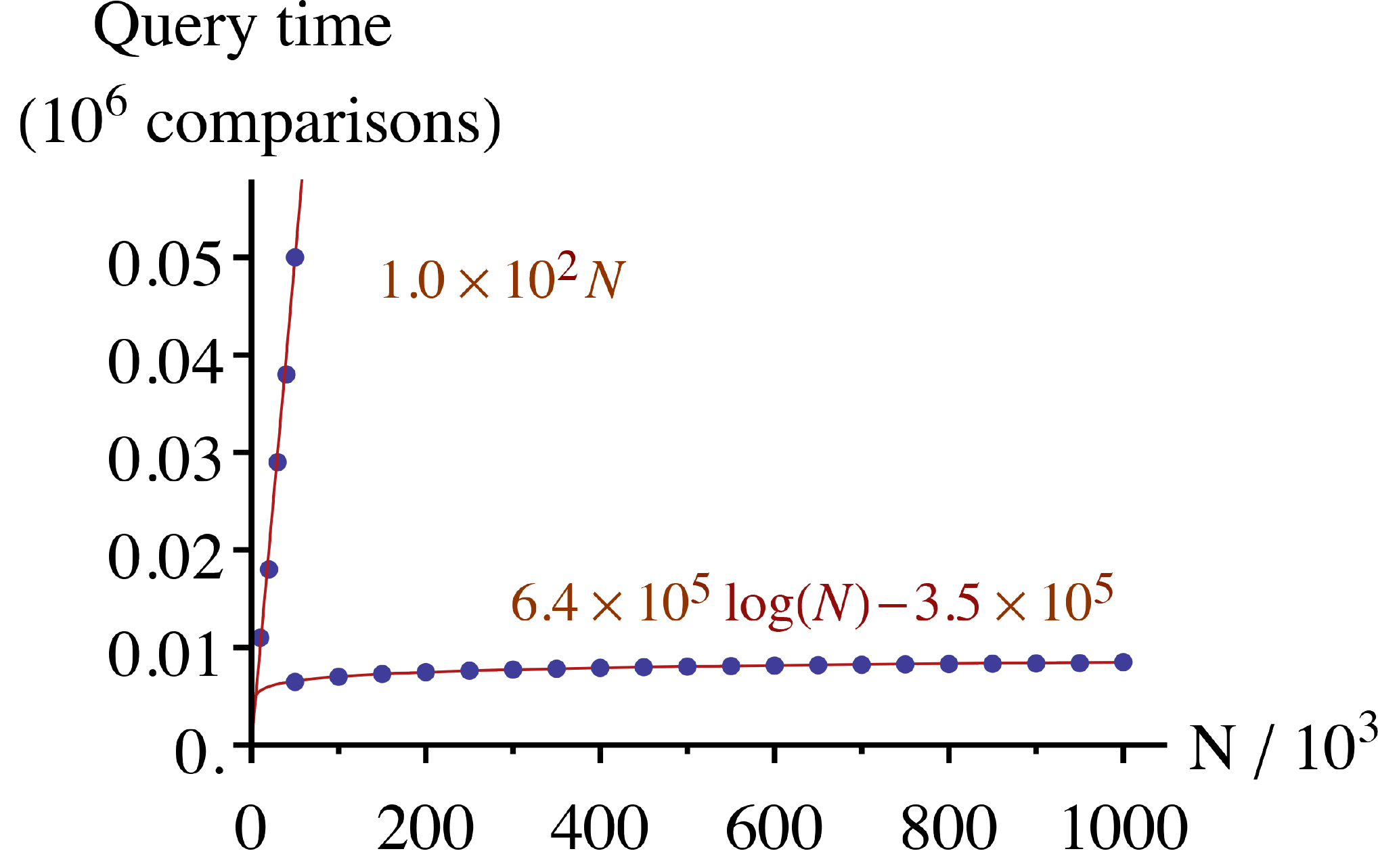}
\caption{
Scaling of (a) training time and (b) query time total number of metric comparisons, for a BF with $n_T=50$ and $k=50$ and a CT with $\epsilon=10$ and $b=1.3$, for uniform random data in a 100-dimensional hypercube. The top line represents the CT, and the bottom line the BF. The CT lines were fit to a quadratic $a N^2 + b N + c$ in (a) and a linear line $a N$ in (b) respectively, while the BF lines fit to $a(N log(N) - N)$ in (a) and $a log(N) + b$ in (b). For all lines the fits were obtained using data up to $N=10^6$. }
\label{fig:BFvsCT}
\end{figure}

In Fig. \ref{fig:BFvsCT} we train a BF with $n_T=50$ and $k=50$, and a CT with $\epsilon=10$ and $b=1.3$ on uniform random data drawn from the 100-dimensional hypercube. We find for this example that training scales quadratically, and querying linearly with the number of points $N$ for the $CT$, while they scale as $ N log(N)$ and $log(N)$ for the BF as seen before. While for the BF training time is the integral over query time, for $CT$ insertion and querying are different. We find that the CT scaling is worse than the BF scaling.


\section{Numerical results} \label{sec:num}
 
The main claim we substantiate in this section is that the BF as a classification or regression algorithm has accuracy comparable to the
$K$-nearest neighbors ($K$-NN) algorithm on real datasets, with a fraction of the computational time, while maintaining the desirable property of learning incrementally.  Since the traversal of the BF is dictated by the metric, the algorithm relies on metric comparisons being informative. Thus, if certain features are much more important than others, BF, like other metric-based methods will perform poorly, unless one can identify or learn a good metric. 

We compare to the highly optimized FLANN \cite{muja2009fast} implementation of multiple random kd trees (R-kd). This algorithm gave the best performance of the ones available in FLANN. We found 4 kd trees gave the best results. One has to set an upper limit to the number of points the kd trees are allowed to visit, which we set to $10\%$ of the training points, a number which led the R-kd to give good performance compared to $1-NN$. Note that R-kd is not parallelizable: the results from each tree at each step inform how to move in the others. 


The datasets we discuss in this section are available at the LIBSVM\cite{libsvm} repository.
Note that for MNIST, we use a permutation-invariant metric based on raw pixel intensities (for easy comparison with other algorithms) even though
other metrics could be devised which give better generalization performance.\footnote{A simple HOG metric gives the BF a $1.1\%$ error rate on MNIST.}
For the BF we set $n_T=50$, $k=50$ for all experiments, and for the RF we use $50$ trees and $50$ features (see Appendix for other choices of parameters). 
We use a laptop with a 2.3 GHz Intel I7 CPU with 16GB RAM running Mac OS 10.8.5.

We find that the BF has similar accuracy to $k-NN$ with a computational cost that scales better, and also the BF is faster than the cover tree, and faster to query than randomized kd trees in most cases (for a CT with $\epsilon=0.1$ shown in appendix, CT becomes faster to train but query time becomes even slower). The results for classification are shown in tables \ref{tab:digitclasstraintest} and \ref{tab:digitclasserr}.

We have also studied the regret, i.e. how much accuracy one loses being online over being offline. In the offline BF each tree gets an independently reshuffled version of the data. Regret is small for all data sets tested, less than $10\%$ of the error rate.

\begin{table}[h!]
\centering
\begin{tabular}{ lrrrrr }
    \put(-20,0){(a)}
\small \quad Data 	& \small	BF 	& \small	BF-4	&\small	R-kd & \small	CT & \small RF \\
  \hline

\small dna 		& 	\small 0.34	&\small	0.15 &	\small 0.042				&\small	0.32	 & \small 3.64	\\
\small letter	 		&\small	1.16	&\small	0.80	&\small	0.12				&\small	1.37	& \small 7.5	\\	
\small mnist &\small	103.9&\small	37.1 &\small		5.67				& \small	168.4	& \small 310	\\
\small pendigits		&\small	0.34	&\small	0.42  &\small		0.059			&\small	0.004 &\small 4.7		\\
\small protein		&\small	35.47 &\small	13.81 &\small	0.90				&\small	44.4	 & \small 191	\\
\small seismic		&\small	48.59	 &\small	16.30	&\small	1.86		&\small	176.1 & \small 2830		\\
  \end{tabular}
  \begin{tabular}{  rrrrrrr }
    \put(-20,0){(b)}
\small  \small \quad	BF 	& \small	BF-4	& \small 1-NN & \small 3-NN &   \small	R-kd & \small	CT & \small RF \\
  \hline
\small \small	0.34	&\small	0.15	 & \small 3.75 & \small 4.23	&\small	0.050		&\small	0.25 & \small 0.025				\\
\small \small	1.16	&\small	0.80 	 & \small 5.5 & \small 6.4		&\small    	1.67 		&\small	0.91	& \small 0.11			\\
\small \small	23.9&\small	8.7 		 & \small 2900 & \small 3200	& \small 	89.2 		&\small	417.6 & \small 0.3				\\
\small \small	0.34	&\small	0.42 		 & \small 2.1 & \small 2.4	&\small  	0.75 		&\small	0.022 & \small 0.03				\\
\small \small	35.47 &\small	13.8 		 & \small 380 & \small 404	&\small  	11.5 		& \small	51.4	 & \small 625			\\
\small \small	16.20 &\small	5.2		 & \small 433 & \small 485	&\small 	65.7		& \small	172.5 & \small 1.32			
  \end{tabular}

    \caption{ (a) Total training time and (b) total testing time, in seconds, for classification benchmarks, single core. In (b) the datasets are in the same order as in (a).
    BF has $n_T=50$, $k=50$.
    For $1-NN$, $3-NN$ and $RF$ we use the Weka\cite{hall2009weka} implementation. 
  RF has 50 trees, 50 features per node. BF-4 is the same BF with 4 cores. Rkd has 4 kd trees and can visit at most $10\%$ of previously seen examples, and points are added online. $CT$ has $\epsilon=10$ and $b=1.3$, and uses online insertion. See Appendix for 10-NN, RF with 100 trees and $\sqrt{D}$ features per node (recommended in \cite{breiman2001}), and $CT$ with $b=1.1$, $\epsilon=0.1$. The datasets are from the LIBSVM repository\cite{libsvm}.}
  \label{tab:digitclasstraintest}
\end{table}

\begin{table}[h!]
\centering
\begin{tabular*}{1.0\columnwidth}{ p{0.8cm} p{0.4cm} r r r r r r }
\small Data 			& \small BF 				& \small OBF		& \small 1-NN		 & \small 3-NN 		& \small RF		&	\small R-kd & \small CT \\
\hline
\small dna				&\small	$14.3$			&\small $13.1$		&\small	$25.0$	&\small	$23.9$	&\small $5.7$  	&\small	$22.5$			&\small $25.55$				\\
\small letter				&\small	$5.4$			&\small $5.5$		&\small	$5.5$ 	&\small	$5.4$	&\small $7.6$	&\small	$5.5$			&\small $5.6$								\\
\small mnist 			&\small	$2.24$   			&\small $2.6$		&\small	$3.08$	&\small      $2.8$ 	&\small $3.2$	&\small	$3.08$			&\small $2.99$					\\
\small pendigits			&\small	$2.62$			&\small $2.60$		&\small	 $2.26$	&\small	$2.2$	&\small $5.2$	&\small	$2.26$			&\small $2.8$					\\
\small protein			&\small	$44.2$			&\small $41.7$		&\small	$52.7$	&\small	$50.7$   &\small $32.8$	&\small	$53.6$			&\small $52.0$					\\
\small seismic			&\small	$40.6$			&\small $ 39.6$		&\small	$34.6$	&\small	$30.7$	&\small $23.7$	&\small	$30.8$			&\small $38.9$
  \end{tabular*}
  \caption{Error rate for classification benchmarks. The values represent percentages.
  The offline BF algorithm does only marginally better on average than the BF. The random kd trees and cover tree are about the same accuracy as $1-NN$. 
  }
  \label{tab:digitclasserr}
\end{table}

The training and testing times for the classification benchmarks are shown in Table \ref{tab:digitclasstraintest}, and the error rates in Table \ref{tab:digitclasserr}. For more results,
We find that indeed BF has similar error rates to $k$-NN, and the sum of training and testing time is a fraction of that for naive $k
$-NN. We emphasize that the main advantage of BFs is the ability to quickly train on and respond to arbitrarily large numbers
of examples (because of logarithmic scaling) as would be obtained in an online streaming scenario. To our knowledge, these properties are unique to BFs as compared with other approximate nearest neighbor schemes.

We also find that for some datasets the offline Random Forest classifier has a higher error rate, and the total training and testing time is higher. Note also that the offline Random Forest needs to be retrained fully if we change the amount of data. On the other hand, there are several data sets for which RFs out-perform BFs, namely those for which it is possible to identify informative sub-sets
of features. Furthermore, we generally find that training is faster for BFs than RFs because BFs do not have to solve a complicated optimization problem, 
but at test time RFs are faster than BFs because computing answers to a small number of decision tree questions is faster than computing distances. 
On the other hand, online R-kd is faster to train since it only does single feature comparisons at each node in the trees, however since it uses less informative decisions than metric comparisons it ends up searching a large portion of the previously seen data points, which makes it slower to test. Note that our main point was to compare to algorithms that do metric comparisons, but these comparisons are informative as well.

\section{Conclusion and future work} \label{sec:future}

We have described and studied a novel online learning algorithm with empirical $ N log(N)$ training and $log(N)$ querying scaling with the amount of data $N$, and similar performance to $k-NN$.

The speed of this algorithm makes it appropriate for applications such as real-time machine learning, and metric learning\cite{weinberger2006distance}. Interesting future avenues would include: combining the BF with random projections, analyzing speedup and impact on performance; testing a real time tracking scenario, possibly first passing the raw pixels through a feature extractor.

\newpage
\bibliographystyle{aaai}
\bibliography{ref}

\newpage

\renewcommand{\thetable}{A-\arabic{table}}
\setcounter{table}{0}  
\setcounter{secnumdepth}{0}  
  \renewcommand{\theequation}{A-\arabic{equation}}
  \setcounter{equation}{0}  

\title{Supplementary Information to ``The Boundary Forest Algorithm for online supervised and unsupervised learning"}
\section*{Supplementary Information to ``The Boundary Forest Algorithm for online supervised and unsupervised learning"}
\author{Charles Mathy\\
Disney Research Boston\\
cmathy@disneyresearch.com\\
\And
Nate Derbinsky\\
Wentworth Institute of Technology\\
derbinskyn@wit.edu\\
\And
Jos\'e Bento\\
Boston College\\
jose.bento@bc.edu\\
\And
Jonathan Yedidia\\
Disney Research Boston\\
yedidia@disneyresearch.com\\
}
\maketitle


\section{Datasets information}
\label{app:data}

\begin{table}[h!]
\centering
\begin{tabular}{ lrrrr }
\small \quad Data 	& \small	$N_C$ 	& \small	$D$	&\small	$N_{train}$ & \small	$N_{test}$ \\
  \hline
\small dna 			& \small 	3	&\small	180 	&	\small 1,400				&\small	1,186		\\
\small letter	 		&\small	26	&\small	16	&\small	10,500				&\small	5,000		\\	
\small mnist 			&\small	10	&\small	784 	&\small		60,00				& \small	168.4		\\
\small pendigits			&\small	10	&\small	16  &\small		7,494			&\small	3,498		\\
\small protein			&\small	3 &\small	357 &\small	14,895				&\small	6,621		\\
\small seismic			&\small	3	 &\small	50	&\small	78,823		&\small	19,705		\\
  \end{tabular}
    \caption{ Information about the datasets in the main text. $N_C$ is the number of classes, $D$ the number of features, $N_{train}$ the number of training examples, and $N_{test}$ the number of test examples. All come from the LIBSVM repository\cite{libsvm}.}
  \label{tab:digitclasstraintest}
\end{table}

\section{Additional results for $k-NN$ and RF}

 \begin{table}[h!]
\centering
\begin{tabular}{ lrr}
    \put(-20,0){(a)}
\small \quad Data 	& \small	$RF-100$  \\
  \hline
\small dna 		&\small 	1.023		\\
\small letter	 	&\small	6.2			\\	
\small mnist 		&\small	365			\\
\small pendigits		&\small	2.9			\\
\small protein		&\small	48.8	 		\\
\small seismic		&\small	332	 		\\
  \end{tabular}
  \begin{tabular}{  lrr }
    \put(-20,0){(b)}
\small  \quad Data 					& \small	$RF-100$ 	& \small	$10-NN$	\\  
\hline
\small dna 	& \small	0.03	&\small	2.45	 	\\
\small letter	&\small	0.44	&\small	3.9 	 	\\
\small mnist	&\small	0.76	&\small	182 		 \\
\small pendigits	&\small	0.09	&\small	1.8 		\\
\small protein	&\small	0.56 	&\small	217 		 \\
\small seismic	&\small	2.1	&\small	18.4					
  \end{tabular}

    \caption{ (a) Total training time and (b) total testing time, in seconds, for classification benchmarks, single core.    
    $RF-100$ has 100 trees, and number of features per node equal to $[\sqrt{D}]$, i.e. the number of features of the dataset rounded to the closest integer (the value recommended by \cite{breiman2001}.  The datasets are from the LIBSVM repository\cite{libsvm}.}
  \label{tab:digitclasstraintest}
\end{table}

\begin{table}[h!]
\centering
\begin{tabular}{ lrr}
\small Data 			& \small $RF-100$			& \small  $10-NN$		 \\
\hline
\small dna				&\small	$6.3$			&\small $17.3$			\\
\small letter			&\small	$5.4$			&\small $7.54$			\\
\small mnist 			&\small	$3.15$   			&\small $3.26$			\\
\small pendigits			&\small	$3.6$			&\small $2.7$			\\
\small protein			&\small	$32.7$			&\small $46$			\\
\small seismic			&\small	$26.3$			&\small $36$		
  \end{tabular}
  \caption{Error rate for classification benchmarks. The values represent percentages. 
  }
  \label{tab:digitclasserr}
\end{table}

\section{Additional results for CT}

 \begin{table}[h!]
\centering
\begin{tabular}{ lrrr}
    \put(-20,0){(a)}
\small \quad Data 	& \small	$CT(1.1,10)$ & \small	$CT(1.3,0.1)$ & \small	$CT(1.1,0.1)$  \\
  \hline
\small dna 		&\small 	0.08		&0.126	&	0.085	\\
\small letter	 	&\small	0.28		&0.23	&	0.26	\\	
\small mnist 		&\small	18.84	&177.75	&	18.34	\\
\small pendigits		&\small	0.0037	&0.0029	&	0.0029	\\
\small protein		&\small	23.89 	&44.81	&	23.22	\\
\small seismic		&\small	115.22 	&191.09	& 	105.52	\\
  \end{tabular}
\begin{tabular}{ lrrr}
    \put(-20,0){(b)}
\small \quad Data 	& \small	$CT(1.1,10)$ & \small	$CT(1.3,0.1)$ & \small	$CT(1.1,0.1)$  \\
  \hline
\small dna 		&\small 	0.31		&	0.297	&	0.53	\\
\small letter	 	&\small	1.01		&	0.85		&	0.92	\\	
\small mnist 		&\small	369.9	&	430.8	&	398.5\\
\small pendigits		&\small	0.007	&	0.005	&	0.005\\
\small protein		&\small	56.45 	&	57.21	&	73.34\\
\small seismic		&\small	227.8	&	208.6	& 	256.6\\
  \end{tabular}

    \caption{ (a) Total training time and (b) total testing time, in seconds, for classification benchmarks, single core.   $CT(b,\epsilon)$ is a cover tree with base parameter $b$ and approximate nearest neighbor parameter $\epsilon$ (see main text). The datasets are from the LIBSVM repository\cite{libsvm}.}
  \label{tab:digitclasstraintest}
\end{table}

\begin{table}[h!]
\centering
\begin{tabular}{ lrrr}
\small \quad Data 	& \small	$CT(1.1,10)$ & \small	$CT(1.3,0.1)$ & \small	$CT(1.1,0.1)$  \\
  \hline
\small dna 		&\small 	24.87		&25.46	&24.7	\\
\small letter	 	&\small	5.92			&5.44	&5.44	\\	
\small mnist 		&\small	2.97			&3.09	&3.09	\\
\small pendigits		&\small	90.4			&90.39	&90.39	\\
\small protein		&\small	52.23	 	&52.74	&52.74	\\
\small seismic		&\small	38.84	 	&40.75	& 40.76	\\
  \end{tabular}
  \caption{Error rate for classification benchmarks. The values represent percentages. 
  }
  \label{tab:digitclasserr}
\end{table}

\end{document}